\documentclass[runningheads]{llncs}
\usepackage[T1]{fontenc}
\usepackage{graphicx}
\usepackage{color}
\usepackage{url}

\usepackage{booktabs}
\usepackage{wrapfig}
\usepackage{subcaption}
\usepackage{amsmath}
\usepackage{amsfonts}
\usepackage{bbm}
\usepackage{bm}

\begin{document}
\title{Even-Ifs From If-Onlys:  \\ Are the Best Semi-Factual Explanations Found Using Counterfactuals As Guides?}
\titlerunning{Semi-Factual Explanations}
%
\author{Saugat Aryal\inst{1,2}\orcidID{0000-0001-6357-3904} \and
Mark T. Keane\inst{1,2}\orcidID{0000-0001-7630-9598}}
\authorrunning{Aryal \& Keane}
%
\institute{School of Computer Science, University College Dublin, Dublin, Ireland \and
Insight Centre for Data Analytics, Dublin, Ireland \\
\email{saugat.aryal@ucdconnect.ie}, \email{mark.keane@ucd.ie}}
\maketitle              
\begin{abstract}
Recently, counterfactuals using \textit{``if-only''} explanations have become very popular in eXplainable AI (XAI), as they describe which changes to feature-inputs of a black-box AI system result in changes to a (usually negative) decision-outcome. Even more recently, semi-factuals using \textit{``even-if''} explanations have gained more attention. They elucidate the feature-input changes that do \textit{not} change the decision-outcome of the AI system, with a potential to suggest more beneficial recourses.  Some semi-factual methods use counterfactuals to the query-instance to guide semi-factual production (so-called \textit{counterfactual-guided methods}), whereas others do not (so-called \textit{counterfactual-free methods}). In this work, we perform comprehensive tests of 8 semi-factual methods on 7 datasets using 5 key metrics, to determine whether counterfactual guidance is necessary to find the best semi-factuals. The results of these tests suggests not, but rather that computing other aspects of the decision space lead to better semi-factual XAI.

\end{abstract}
%
%
%
\section{Introduction}
The increasing deployment of black-box AI systems in many application domains has increased the need for an eXplainable Artificial Intelligence (XAI). XAI methods aim to surface the inner workings of such models to improve their interpretability, to foster trust, and to audit for responsible use. Recently, counterfactuals have become very popular in XAI; using ``if-only'' explanations they tell end-users about how decision-outcomes can be \textit{altered} when key feature-inputs change \cite{miller2019explanation,keane2021if,karimi2022survey,verma2020counterfactual}.  For instance, if a customer wants a loan refusal explained, they could be told ``if you had only asked for a lower loan, you would have been successful".  
``Even-if'' explanations -- so-called \textit{semi-factuals} -- are closely related to counterfactuals, but differ in that they inform users about how a decision-outcome \textit{stays the same} when key feature-inputs change \cite{aryal2023even,kenny2023sf,kenny2021generating}. For instance, if someone is successful in their loan application, they could be told ``Even if you asked for a longer term on the loan, you would still be successful".  So, like counterfactuals, semi-factuals give people options for algorithmic recourse in automated decision-making \cite{karimi2021algorithmic}.  In this paper, we consider a key question that divides the semi-factual literature; namely, whether the best semi-factuals are to be found by using counterfactuals as guides?

\begin{figure*}[t]
\centering
\begin{subfigure}{.5\textwidth}
  \centering
  \includegraphics[width=.95\linewidth]{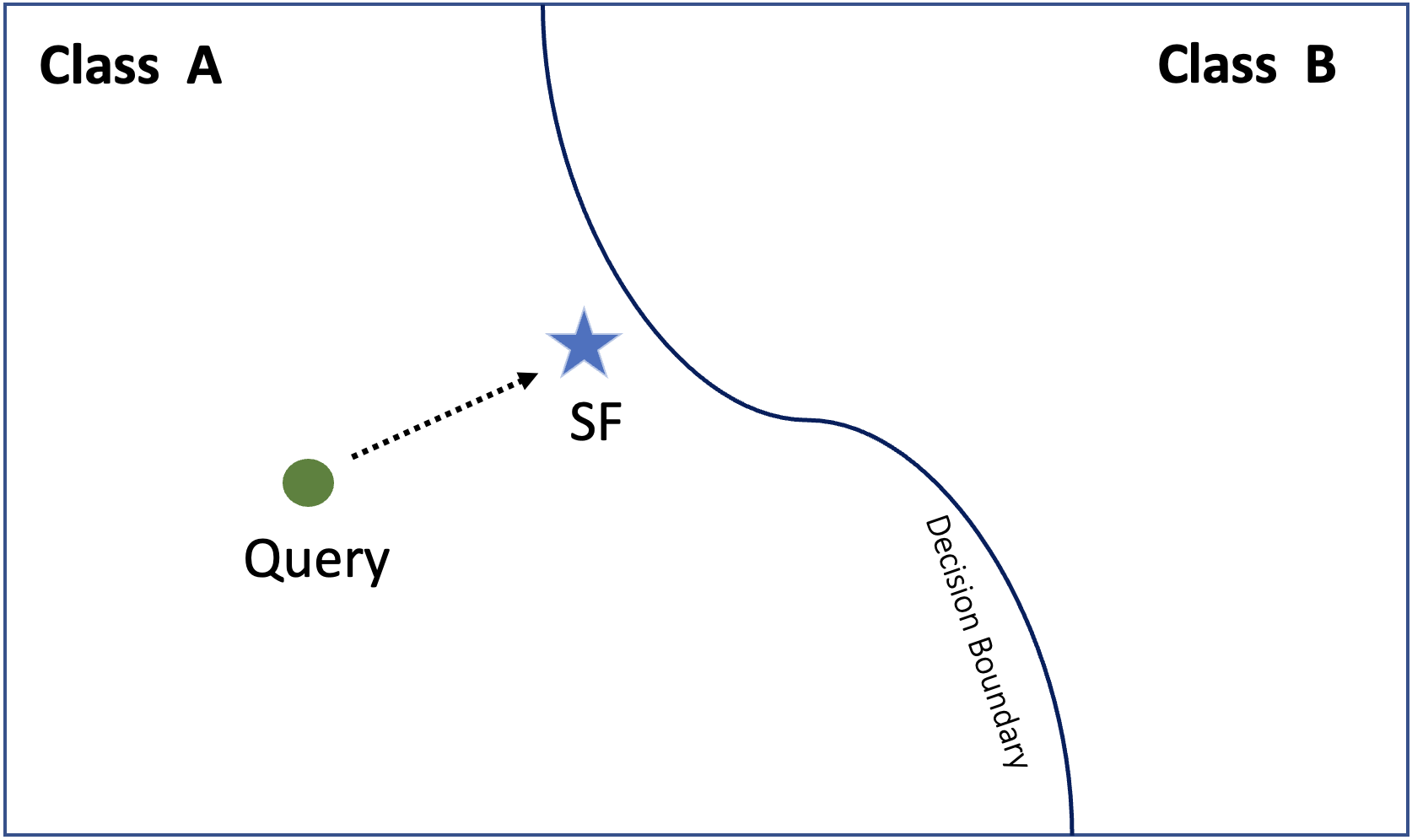}
  \caption{Counterfactual-Free}
  \label{fig:sub1}
\end{subfigure}%
\begin{subfigure}{.5\textwidth}
  \centering
  \includegraphics[width=.95\linewidth]{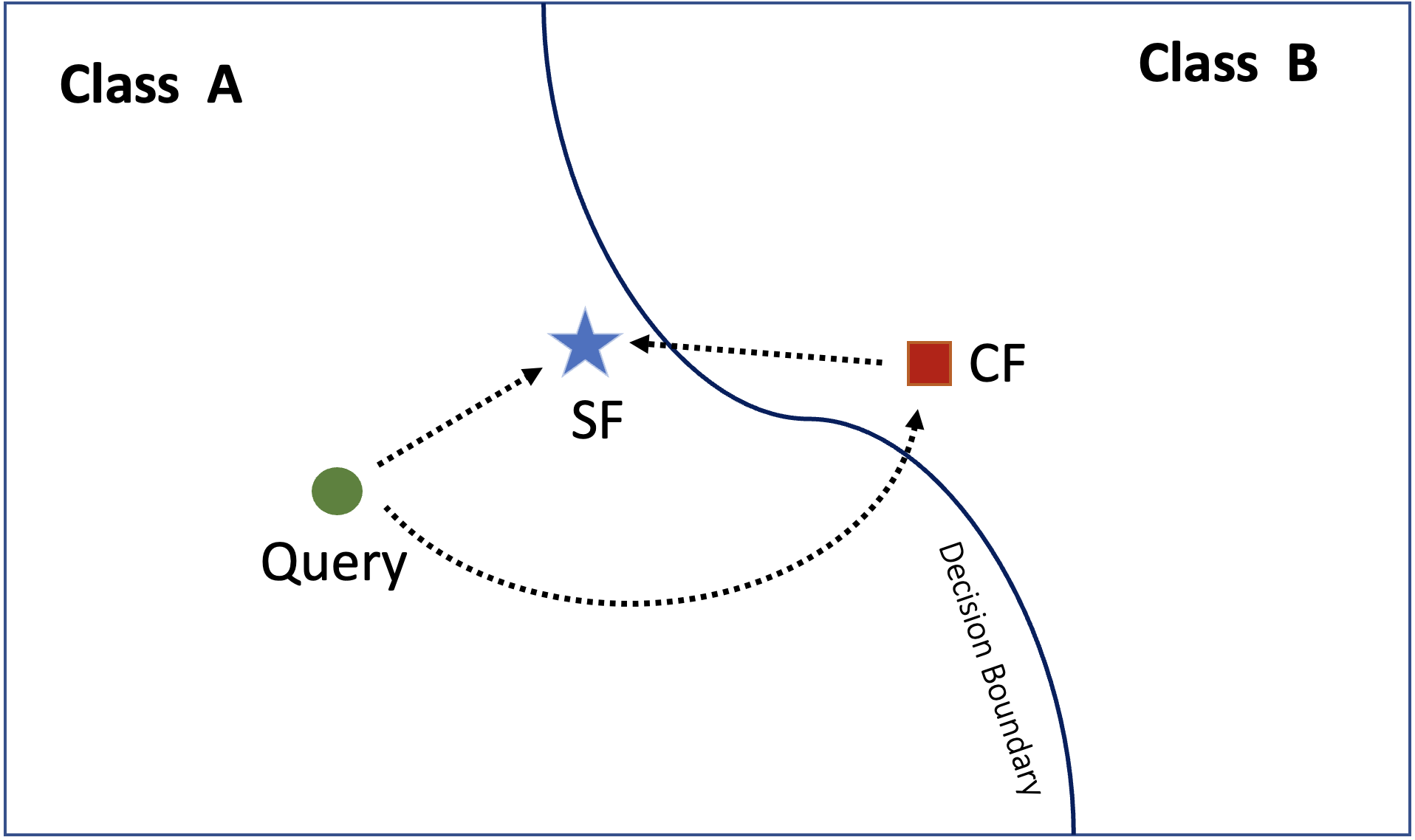}
  \caption{Counterfactual-Guided}
  \label{fig:sub2}
\end{subfigure}
\caption{Visualisation of (a) Counterfactual-Free and (b) Counterfactual-Guided Semi-factual Methods. The semi-factual, \textit{SF}, is found by some computation (a) to be 
within the query-class at some distance from the query, \textit{Q}, or (b) that is guided by the location of a counterfactual, \textit{CF}, for the query, \textit{Q}.}
\label{fig:background}
\end{figure*}

Semi-factual explanations for XAI have received a lot less research attention than countefactual explanations.  Although both methods have strong and long-standing roots in explanatory case-based reasoning (XCBR)  \cite{nugent2005best,cummins2006kleor,nugent2009gaining}, research on counterfactuals exploded around 2019 (for reviews see \cite{miller2019explanation,keane2021if,karimi2022survey,verma2020counterfactual,keane2021if}), whereas semi-factuals were largely passed over (for reviews see \cite{kenny2021generating,aryal2023even,poche2023natural}). Hence, there are now possibly $>$150 distinct XAI methods that compute counterfactual explanations using either optimisation techniques \cite{wachter2017counterfactual,mothilal2020explaining,karimi2021algorithmic}, instance-based approaches \cite{keaneGoodCFandWhereToFindThem,smyth2022few,brughmans2023nice}, genetic algorithms \cite{guidotti2019factual,schleich2021geco}, deep neural networks \cite{hamman2023robust,van2021conditional} and so on.  In contrast, there are perhaps $<$10 distinct methods that compute semi-factuals, many of which were proposed almost 20 years ago in CBR \cite{nugent2005best,cummins2006kleor,nugent2009gaining}.  However, there are reasons to believe that semi-factuals have the potential to be as useful as counterfactuals (c.f., recent work \cite{kenny2023sf,artelt2022even,mertes2022alterfactual,lu2022rationale,vats2022changes,kenny2021generating}).


Interestingly, the motivation for early semi-factual work was to find ``better'' nearest neighbors for case-based explanations, instances that were not \textit{actually} the closest to the query-case but, nevertheless, ones that provided better explanations.  For example, if Mary was refused a loan and Belinda was her nearest neighbor (both are the same age, have same income, same credit score and asked for the same loan), rather than using Belinda to explain the decision we might use Sally (who  differs from Mary in having a much higher credit score); here, the semi-factual explanation is saying``Even if Mary had a very high credit score like Sally, she would still have been refused the loan", providing a stronger or \textit{a fortiori} argument for why Mary was turned down \cite{nugent2005best,nugent2009gaining}. Accordingly, semi-factuals fundamentally owe their existence in XAI to decades of CBR research.

One of the key computational steps in computing semi-factuals appears to be identifying the key feature(s) to change to provide a more convincing explanation.  
Some methods try to achieve this aim heuristically, by finding instances that are as far as possible from the query while still being in the query-class. 
Figure \ref{fig:background}(a) shows this situation graphically with a semi-factual (blue star) that is far from the query (green circle) but still within the query-class (Class-A) and close to the decision boundary with the counterfactual-class (Class-B). Other methods try to find instances that are still in the query-class but which bear some relationship to a counterfactual for the query (e.g., using a nearest unlike neighbour or NUN). 
For instance, in Figure \ref{fig:background}(b), the semi-factual (blue star) is found between the query (green circle) and its counterfactual (red box), using the latter to guide the search.

This opposition reveals two groups of  methods, where one is \textit{counterfactual-free }(i.e., they find semi-factuals without explicitly considering counterfactuals) and the other is \textit{counterfactual-guided} (i.e., they find semi-factuals using counterfactuals as guides). Using counterfactuals makes intuitive sense as they identify key features that flip class-membership, features that might also work well in a semi-factual. However, we do not know whether this counterfactual guidance necessarily or always produces the best semi-factuals.  Hence, in this paper, we consider the key question ``Are the best semi-factuals found by using counterfactuals as guides?"

\textbf{Outline of Paper \& Contributions.} In the remainder of this paper, we first outline the related work on semi-factual methods for XAI, detailing the various methods that have been proposed (see section 2) before presenting our comparison of the performance of 8 different semi-factual methods (4 counterfactual-free and 4 counterfactual-guided methods) across 7 representative datasets on 5 key evaluation metrics (see section 3).  We finish with a discussion of the implications and limitations of our findings (see section 4).   As such, the work makes several novel contributions to this literature, including: 

\begin{itemize}
\item 
A comprehensive analysis of different semi-factual explanation methods divided into counterfactual-free and counterfactual-guided groups.
\item
The most comprehensive set of tests in the literature using key evaluation measures for determining the best semi-factual explanations 
\item 
The discovery of a set of key findings about the dynamics of different semi-factual methods, using/not-using counterfactuals, thus providing baselines and targets for future development in this important area.
\end{itemize}

\section{Semi-Factuals: Counterfactual-Free Or -Guided}

Technically speaking, semi-factuals are a special case of the counterfactual, in that they assert facts that are ``counter'' to the reality of what actually happened, albeit facts that do not change outcomes, as opposed to facts that change outcomes (as in counterfactuals \textit{proper}). However, in philosophy it has been argued that semi-factuals are fundamentally different to counterfactuals \cite{bennett2003philosophical} and, psychologically, they cognitively impact people differently \cite{mccloy2002semifactual}; McCloy \& Byrne \cite{mccloy2002semifactual} found that people judge an antecedent event to be more causally-related to the outcome when given counterfactuals, but judge antecedents to be less causally-related to the outcomes when given semi-factuals.  Aryal \& Keane \cite{aryal2023even} reviewed the history of semi-factual XAI research from the 2000s to 2023, but failed to note that groups of methods differed in their reliance on counterfactuals (i.e., counterfactual-guided v counterfactual-free). Interestingly, these different approaches are not historically-correlated, older and newer methods are equally likely in both groups.  So, some researchers consider counterfactuals to be essential in computing semi-factuals whereas others do not. Here we explore these two approaches, to ascertain whether counterfactual guidance is required to compute the best semi-factuals.   Hence, in the following sub-sections we quickly profile eight semi-factual methods (4 counterfactual-free and 4 counterfactual-guided methods)  tested here.

\subsection{Counterfactual-Free Methods}
Counterfactual-free methods tend to work within the query-class exploring the relationship between candidate semi-factuals and the query (e.g., using distance) but are all quite distinct in how they work.  Historically ordered they are the (i) Local-Region Model \cite{nugent2009gaining}, (ii) DSER \cite{artelt2022even}, (iii) MDN \cite{aryal2023even}, and (iv) S-GEN \cite{kenny2023sf}.

\subsubsection{Local-Region Model.}
Nugent et al. [2009] proposed the Local-Region model, which analyses the local region around the query, using a surrogate model (akin to way  LIME works \cite{ribeiro2016should}), to select the nearest neighbor in the query class with the most marginal probability, as the semi-factual instance; as follows:

\begin{equation}
Local\text-Region(q, C)=\arg \min_{x \epsilon C} LR(x)
\end{equation}

\noindent where, \emph{C} is the set of candidate neighbors and \emph{LR()} is the local logistic regression model providing the probability score. Even though this method considers instances from non-query classes (e.g., the counterfactual class) to plot the decision boundary, we consider it to be counterfactual-free as it does not single out specific counterfactual-instances to guide the process of semi-factual selection.

\subsubsection{Diverse Semifactual Explanations of Reject (DSER).} 
Artelt \& Hammer [2022] proposed a semi-factual method to explain reject decisions in machine learning; that is, to explain why a model should \textit{not} make a prediction. Conceptually, DSER is patterned on the optimisation methods proposed for counterfactuals, though obviously uses somewhat different constraints (see e.g., \cite{wachter2017counterfactual,mothilal2020explaining,dandl2020multi}); it applies its loss function to candidate-instances in the query-class, using four constraints for good semi-factuals (i.e., feasibility, sparsity, similarity, diversity):

\begin{equation}
\textit{DSER}(q)=\arg \min_{q_{sf} \epsilon \mathbb{R}^d} \ell(q_{sf})
\end{equation}

\noindent where, $q_{sf}$ is the semi-factual of query $q$ and $\ell()$ represents the combined loss function such that, 

\begin{equation}
    \ell(q_{sf})=\ell_{feasibile}(q_{sf})+\ell_{sparse}(q_{sf})+\ell_{similar}(q_{sf})+\ell_{diverse}(q_{sf})
\end{equation}

\noindent where feasibility is cast as,

\begin{equation}
\ell_{\text {feasible }}\left(q_{sf}\right) =C_{\text {feasible}} \cdot \max \left(r\left(q_{sf}\right)-\theta, 0\right)+C_{sf} \cdot \max \left(r(q)-r\left(q_{sf}\right), 0\right)
\end{equation}

\noindent which ensures that the semi-factual is also predictively uncertain but more certain than the original query, \emph{q} (to be convincing). Here, $C$ represents the regularization parameters for each component, $r()$ is the reject function based on the certainty of predictive function and $\theta$ is the reject threshold. 

\begin{equation}
\ell_{\text {sparse}}\left(q_{sf}\right)=C_{\text {sparse}} \cdot \max \left(\sum_{i=1}^d \mathbbm{1}\left(\left(q_{sf}-q\right)_i \neq 0\right)-\mu, 0\right)
\end{equation}

\noindent covers \textit{sparsity} promoting candidates with fewer feature differences between the semi-factual and the query. Here $d$ is the number of feature dimensions and $\mu \geq 1$ is a hyperparameter that controls the number of feature-differences.  

\begin{equation}
\ell_{\text {similar}}\left(q_{sf}\right)=-C_{\text {similar }} \cdot\left\|q_{sf}-q\right\|_2
\end{equation}

\noindent deals with \textit{similarity} promoting greater distance between the query and the semi-factual in Euclidean space, and finally,

\begin{equation}
\ell_{\text {diverse }}\left(q_{sf}\right)=C_{\text {diverse }} \cdot \sum_{j \in \mathcal{F}} \mathbbm{1}\left(\left(q_{sf}-q\right) \neq 0\right)
\end{equation}

\noindent handles \textit{diversity} ensuring that several featurally-distinct semi-factuals are generated. Here, $\mathcal{F}$ represents the set of features that have already been used to generate semi-factuals, feature-sets that should be avoided:

\begin{equation}
\mathcal{F}=\left\{j \mid \exists i:\left(q_{sf}^i-q\right)_j \neq 0\right\}
\end{equation}

\subsubsection{Most Distant Neighbor (MDN).}
Aryal \& Keane [2023] proposed and tested this method as a na\"ive, benchmark algorithm to find query-class instances that are most distant from the query. MDN scores all the query-class's instances on the extremity of their feature-values, determining whether they are much higher or lower than the feature-values of the query, \textit{q}, to find its most distant neighbor. Its custom distance function, \textit{Semi-Factual Scoring} (\textit{sfs}), prioritises instances that are sparse relative to the query (i.e., fewer feature differences), but have the highest value-differences in their non-matching features, as follows:

\begin{equation}\label{eq:sfs}
\textit{sfs}(q, S, F)=\frac{same(q, x)}{F}+\frac{\textit{diff}(q_f, x_f)} {\textit{diff}_{max}(q_f, S_f)}
\end{equation}

\noindent where \textit{S} is Higher/Lower Set and $x \in S$, \textit{same()} counts the number of features that are equal between \textit{q} and \textit{x}, \textit{F} is the total number of features, \textit{diff()} gives the difference-value of key-feature, \textit{f}, and \textit{diff}$_{max}()$ is the maximum difference-value for that key-feature in the Higher/Lower Set. The best-feature-MDN is selected as the instance with the highest \textit{sfs} score from the Higher/Lower set for each feature, independently. Finally, the best of the best-feature-MDNs across all dimensions is chosen as the overall semi-factual for the query,

\begin{equation}
\textit{MDN}(q, S)=\arg \max_{x \in S} sfs(x)
\end{equation}

\noindent MDN finds the query-class instance that is furthest away from query on some dimension(s), one that also shares many common features with the query. As such, MDN never considers instances from other classes (such as, a counterfactual class) when it processes candidate semi-factuals. In the current tests, we use an improved version of MDN based on a new scoring function, \textit{sfs}$_{v2}$, that assigns a greater weight to the sparsity component, as follows:

\begin{equation}\label{eq:sfs_v2}
\textit{sfs}_{v2}(q, S, F)=\frac{1}{F - same(q,x)}*(\frac{same(q, x)}{F}+\frac{\textit{diff}(q_f, x_f)} {\textit{diff}_{max}(q_f, S_f)})
\end{equation}

\noindent as the scoring function in Eq. (\ref{eq:sfs}) performs poorly on sparsity  (see \cite{aryal2023even} for tests).

\subsubsection{Explanations for Positive Outcomes (S-GEN).}
In 2023, Kenny \& Huang \cite{kenny2023sf} proposed the novel concept of ``gain" (akin to ``cost" in counterfactuals \cite{ustun2019actionable}) as a new constraint for semi-factual methods.   They argue that semi-factuals best explain positive outcomes, whereas counterfactuals work best for negative outcomes. For example, if I am granted a loan at a 5\% interest-rate, a comparison-shopping semi-factual might suggest that I could be granted the loan at a 2\% interest-rate (from another bank). This semi-factual explanation has a quantifiable gain for end users, that can also be used to select explanations.  Therefore, S-GEN uses \textit{gain}, along with traditional constraints (such as plausibility, robustness and diversity) to compute semi-factuals that tell users about better, positive recourses that could be used. S-GEN's objective function is : 
\nopagebreak
\begin{equation}\label{eq:sfpositive}
\begin{aligned}
S\text-GEN(\mathbf{q})=\max _{\mathbf{a}_1, \ldots, \mathbf{a}_m} & \frac{1}{m} \sum_{i=1}^m f\left(P\left(\mathbf{q}, \mathbf{a}_i\right), G\left(\mathbf{q}, \mathbf{a}_i\right)\right)+\gamma R\left(\left\{\mathbf{q}_1^{\prime}, \ldots, \mathbf{q}_m^{\prime}\right\}\right) \\
\text { s.t. } & \forall i, j: \mathbf{q}_i^{\prime}=S_{\mathcal{M}}\left(\mathbf{q}, \mathbf{a}_i\right), H_j\left(\mathbf{q}_i^{\prime}\right) \geq 0(\text { or }>)
\end{aligned}
\end{equation}

\noindent where, $\mathbf{a_i}$ represents an action taken on $i^{th}$ feature-dimension, $m$ is the desired number of semi-factuals to be generated. $
P(\mathbf{q}, \mathbf{a}_i)=\operatorname{Pr}\left(S_{\mathcal{M}}(\mathbf{q}, \mathbf{a}_i)\right)
$ denotes the plausibility of explanation for $\mathbf{q}$ by taking action $\mathbf{a}_i$ where $Pr$ is the distribution density. S-GEN uses a Structural Causal Model (SCM), $S_{\mathcal{M}}$ to better capture the causal dependencies between the features and hence obtain feasible explanations.

\begin{equation}
G(\mathbf{q}, \mathbf{a}_i)=\mathcal{P}_{SF} \circ \delta\left(\mathbf{q}, S_{\mathcal{M}}(\mathbf{q}, \mathbf{a}_i)\right)
\end{equation}

\noindent represents the gain function for $\mathbf{q}$ by taking action $\mathbf{a}_i$ where $\delta()$ is the distance function. Intuitively, it measures the difference between original state $\mathbf{q}$ and the new state obtained by the transition from $\mathbf{q}$ by taking action $\mathbf{a}_i$ through an SCM, $S_{\mathcal{M}}(\mathbf{q},\mathbf{a}_i)$. Greater differences indicate higher ``gains'' (i.e., better explanations).

\begin{equation}
R\left(\left\{\mathbf{q}_i\right\}_{i=1}^m\right)=\frac{2}{m(m-1)} \sum_{i=1}^m \sum_{j>i}^m L_p \circ \delta\left(\mathbf{q}_i, \mathbf{q}_j\right)
\end{equation}

\noindent shows the diversity function which is regularized by $\gamma$ in Eq. (\ref{eq:sfpositive}).  $L_p$ is the \mbox{$L_p$-norm} and $\delta()$ is the distance function used, so as many distinct explanations as possible are generated that are also far from each other. Finally,  robustness is achieved in a post-hoc manner as a hard constraint defined by:

\begin{equation}
H(\mathbf{q}, \mathbf{a})=\min _{\mathbf{q}^{\prime} \in \mathbb{B}_s(\mathbf{q}, \mathbf{a})} h\left(\mathbf{q}^{\prime}\right)-\psi
\end{equation}

\noindent where $H(\mathbf{q}, \mathbf{a})$ denotes the post-robustness of an action $\mathbf{a}$ for a test instance $\mathbf{q}$. The intuition is that any instances lying in the neighborhood $\mathbb{B}_s$ of the generated semi-factual $\mathbf{q}^\prime=S_{\mathcal{M}}(\mathbf{q},\mathbf{a})$ after taking the action $\mathbf{a}$ also have a positive outcome. This function ensures that the output of a predictive model $h$ for $\mathbf{q}^\prime$ is higher than a threshold $\psi=0.5$ (in case of binary classes).

\subsection{Counterfactual-Guided Methods}
Both counterfactual-free and counterfactual-guided methods try to find instances that are distant from the query to use as semi-factuals.  However, counterfactual-guided methods differ in their use of counterfactuals to the query as indicators/guides to good semi-factuals.  The four methods examined are (i) KLEOR~\cite{cummins2006kleor}, (ii) PIECE \cite{kenny2021generating}, (iii) C2C-VAE \cite{zhao2022generating}, and (iv) DiCE \cite{mothilal2020explaining}. 
\vspace{-2mm}
\subsubsection{Knowledge-Light Explanation-Oriented Retrieval (KLEOR).} 
Cummins \& Bridge [2006] were the first to use counterfactuals to guide the selection of a semi-factual. They proposed several methods, using different distance measures, that identify query-class instances as semi-factuals, when they lie between the query and its counterfactual (aka its NUN); viewing the best semi-factual as that one closest to the NUN and furthest from the query. Their \emph{Sim-Miss} method: 

\begin{equation}
Sim\text-Miss(q, nun, G) = \arg \max_{x \epsilon G} Sim(x, nun)
\end{equation}

\noindent where \textit{q} is the query, \textit{x} is a candidate instance, $G$ is the set of query-class instances, and \textit{nun} is the NUN, and \textit{Sim} is Euclidean Distance.  \emph{Global-Sim}, has the semi-factual lie between \emph{q} and the \emph{nun} in the overall feature space:
\vspace{-1mm}
\begin{equation}
Global\text-Sim(q, nun, G) = \arg \max_{x \epsilon G} Sim(x,nun) + Sim(q, x) > Sim(q, nun)
\end{equation}

\noindent whereas the third, \emph{Attr-Sim}, enforces similarity  across a majority of features:
\begin{equation}
\begin{split}
{Attr\text-Sim}(q, nun, G) &= \arg \max_{x \epsilon G} Sim(x,nun) \\
&\qquad {}+ \max_{a \epsilon F} count[Sim(q_a, x_a) > Sim(q_a, nun_a)]
\end{split}
\end{equation}

\subsubsection{PlausIble Exceptionality-based Contrastive Explanations (PIECE).}
Kenny \& Keane's \cite{kenny2021generating} PIECE method computes semi-factuals ``on the way to'' computing counterfactuals using statistical techniques and a Generative Adverserial Network (GAN) model, that can work with both image and tabular data. PIECE identifies ``exceptional'' features in a query with respect to its counterfactual class (i.e., probabilistically-low in that class) and, then, iteratively modifies these features in the query until they are ``normal'' (i.e., probabilistically-high). As these exceptional features are incrementally altered, the generated instances gradually move away from the query towards the counterfactual class, with the last instance just before the decision boundary being deemed to be the semi-factual. So, the semi-factual is like a point on the trajectory from the query to the counterfactual.

Exceptional features are identified using the statistical probabilities in the training distribution of the counterfactual class $c^{\prime}$. Specifically, a two-part hurdle process is used to model the latent features of the query (when it is an image) in the feature-extracted layer ($\mathbf{X}$) of a Convolutional Neural Network (CNN) with an ReLU activation function. The first hurdle process is modelled as a Bernoulli distribution and the second as a probability density function (PDF) as: 
\vspace{-1mm}
\begin{equation}
    p\left(x_i\right)=\left(1-\theta_i\right) \delta_{\left(x_i\right)(0)}+\theta_i f_i\left(x_i\right), \quad \text { s.t. } \quad x_i \geq 0
\end{equation}

\noindent where $p(x_i)$ is the probability of the latent feature value $x_i$ for $c^{\prime}$, $\theta_i$ is the probability of the neuron in $\mathbf{X}$ activating for $c^{\prime}$ (initial hurdle process), and $f_i$ is the subsequent PDF modelled (the second hurdle process). The constraint of $x_i \geq 0$ refers to the ReLU activations, and $\delta_{\left(x_i\right)(0)}$ is the Kronecker delta function, returning 0 for $x_i > 0$, and 1 for $x_i = 0$.  After modelling the distribution, a feature value $x_i$ is regarded as an exceptional feature for the query image in situations where, 
\vspace{-1mm}
\begin{equation}
x_i=0 \mid p\left(1-\theta_i\right)<\alpha
\end{equation}

\noindent if the neuron $\mathbf{X}_i$ does not activate, given the probability of it not activating being less than $\alpha$ for $c^{\prime}$, and,

\begin{equation}
x_i>0 \mid p\left(\theta_i\right)<\alpha
\end{equation}

\noindent if a neuron activates, given that the probability
of it activating being less than $\alpha$ for $c^{\prime}$, where $\alpha$ is a threshold.

Once the exceptional features are identified, the query's features are adjusted to their expected values ($x^{\prime}$) with generated instances being checked by the CNN to be in the query or counterfactual-class. The semi-factual is the last generated instance in the query-class before crossing into the counterfactual-class. Finally, a GAN is used to visualize the explanations by identifying a latent vector ($z^{\prime}$) such that loss between $x^{\prime}$ and $C(G(z^{\prime}))$ is minimized as,

\begin{equation}
z^{\prime}=\underset{z}{\arg \min }\left\|C(G(z))-x^{\prime}\right\|_2^2
\end{equation}

\begin{equation}
    \textit{PIECE}(q) = G(z^{\prime})
\end{equation}

\noindent where $C$ is a CNN classifier and $G$ is the GAN generator.

\subsubsection{Class-to-Class Variational Autoencoder (C2C-VAE).}
Ye and colleagues \cite{zhao2022generating,ye2020applying,ye2021learning} have proposed several models based on autoencoders that efficiently compute counterfactuals and semi-factuals. Their C2C-VAE model \cite{zhao2022generating} learns an embedding space representing the differences between feature patterns in two classes using a standard variational autoencoder (VAE), with an encoder $(f)$ and a decoder $(f^{\prime})$. In the initial learning phase, given a pair of cases $s$ and $t$ from two classes, C2C-VAE encodes the feature difference, $f_{\Delta}$, where $f_{\Delta}(s, t)=f(s)-f(t)$ using an encoder $g$, as $g\left(<f_{\Delta}, C_s, C_t>\right)$ and decodes the embedding using a decoder $g^{\prime}$ as $f_{\Delta}^{\prime}=g^{\prime}\left(g\left(<f_{\Delta}, C_s, C_t>\right)\right.$

To derive an explanation for a query, $q$ in class $C_q$, the method first generates a guide $t$ in the counterfactual class $C_t$. This guide selection leverages the feature difference embedding space $g$. Specifically, the method randomly samples vectors from $g$, decodes them back to the original feature space, and selects the one with the least mean squared error compared to $q$. Finally, it interpolates between the extracted features of $f(q)$ and $f(t)$ in the VAE's latent space to obtain counterfactuals and semi-factuals as, 

\begin{equation}
C2C\text-VAE(q)=f^{\prime}((1-\lambda) * f(q)+\lambda * f(t)),  0 \leq \lambda \leq 1
\end{equation}

\noindent where $\lambda$ is a hyperparameter which determines the weight of interpolation between $q$ and $t$ and controls whether the output is more similar to $q$ (for a semi-factual) or $t$ (for a counterfactual). 

\subsubsection{Diverse Counterfactual Explanations (DiCE).} 
Mothilal et. al's \cite{mothilal2020explaining} DiCE is a popular optimisation-based counterfactual method that can produce semi-factuals with adjustments to its loss function. They pioneered constraints for \textit{diversity} (using determinantal point processes) and \textit{feasibility} (using causal information from users) optimizing explanations with the loss function: 
\vspace{-5mm}

\begin{equation}
\begin{split}
DiCE(q)&=\underset{\boldsymbol{c}_1, \ldots, c_k}{\arg \min } \frac{1}{k} \sum_{i=1}^k \operatorname{yloss}\left(f\left(c_i\right), y\right)+\frac{\lambda_1}{k} \sum_{i=1}^k \operatorname{dist}\left(c_i, q\right) \\
&\qquad {}- \lambda_2 \text {dpp\_diversity}(c_1,...,c_k)
\end{split}
\end{equation}

\noindent where $q$ is the query input, $c_i$ is a counterfactual explanation, $k$ is the total number of diverse counterfactuals to be generated, $f()$ is the black box ML model, $yloss()$ is the metric that minimizes the distance between $f()$'s  prediction for $c_i$ and the desired outcome $y$, $dist()$ is the distance measure between $c_i$ and $q$, and dpp\_diversity() is the diversity metric. $\lambda_1$ and $\lambda_2$ are hyperparameters that balance the three components of the loss function.

\section{Experimental Evaluations: To Guide or Not to Guide?}

Taking these 8 semi-factual methods we performed a series of  evaluations using 7 commonly-used datasets to determine whether the best semi-factuals are found by using counterfactuals as guides.  We pitted the counterfactual-free methods against the counterfactual-guided ones, assessing them on 5 key metrics: distance, plausibility, confusability, robustness and sparsity.  These tests are the most comprehensive experiments to date in the semi-factual XAI literature, in the number of methods/datasets tested and the metrics used. 

\subsection{Method: Metrics}
Our computational evaluations of semi-factual methods used the key metrics that capture aspects of the desiderata for ``good" semi-factuals (see \cite{aryal2023even}):

\textbf{Distance.} It is commonly-held that good semi-factuals are further from the query; here, measured as $L_2$-norm distance, where higher is better.

\textbf{Plausibility.} The semi-factual should be plausible; that is, within-distribution; here, measured as the $L_2$ distance to the nearest training instance, where lower values are better (see also \cite{kenny2023sf}).

\textbf{Confusability.} The semi-factual should not be so far from the query so as to be confusable with a counterfactual. That is, the classifier should have higher confidence in classifying in the query class rather than in the counterfactual class; here, measured as ratio of its distance to the core of the counterfactual and query classes (adapted from Jiang et al's \cite{jiang2018trust} Trust Score). This score is reversed (using $1-score$) so lower values are better (i.e., less confusable).

\begin{equation}\label{eq:confusability}
\textit{Confusability(x)}=\frac{\textit{d(x,CF)}}{\textit{d(x,Q)}}
\end{equation}
\noindent where $x$ is the semi-factual, \textit{CF} is the counterfactual-class, $Q$ is the query-class and $d()$ measures the distance.

\textbf{Robustness.} The semi-factual method should be robust, small perturbations of the query should not radically change the semi-factuals found; here, measured using local Lipschitz continuity (see \cite{alvarez2018robustness}). Specifically, we calculate, 

\begin{equation}\label{eq:robustness}
Robustness(x)=\underset{x_i \in B_\epsilon\left(x\right)}{\operatorname{argmax}} \frac{\left\|f\left(x\right)-f\left(x_i\right)\right\|_2}{\left\|x-x_i\right\|_2}
\end{equation}

\noindent where $x$ is the input query, $B_\epsilon\left(x\right)$ is the ball of radius $\epsilon$ centered at $x$, $x_i$ is a perturbed instance of $x$ and $f()$ is the explanation method. Again this score was reversed so higher values are better (i.e., more robust). 

\textbf{Sparsity.} There should be few feature differences between the query and semi-factual to be easily comprehended \cite{keaneGoodCFandWhereToFindThem}; here, measured using,
\begin{equation}
Sparsity  = \frac{ideal_{\textit{f\_diff}}}{observed_{\textit{f\_diff}}}
\end{equation}
\noindent where $ideal_{\textit{diff}}$ is the desired number of feature differences (we use 1) and $observed_{\textit{diff}}$ is the number of differences present, a higher value being better.

\subsection{Method: Datasets \& Setup}
The eight methods were assessed with respect to each evaluation measure across seven benchmark tabular datasets commonly used in the XAI; namely, Adult Income (D1), Blood Alcohol (D2), Default Credit Card (D3), Diabetes (D4), German Credit (D5), HELOC (D6) and Lending Club (D7). For a thorough examination, 5-fold cross-validation was used to evaluate each method on each dataset. We set $\epsilon = 0.1$ in Eq. (\ref{eq:robustness}) and obtain 100 perturbations within that radius. In case of PIECE, DiCE, and S-GEN, feasibility constraints were added for the datasets allowing only certain features to mutate (see GitHub repository for details).

The Attr-Sim variant was implemented for KLEOR as it is the most sophisticated variant of this method and used a 3-NN model to build it. PIECE was adapted to accept tabular data by fitting the training data using a gamma distribution and calculating the probability of expected value through cumulative distribution function (CDF). We used feed-forward layers to train the VAE and C2C-VAE models in the PyTorch framework (see details about the architecture and hyperparameters in GitHub). The value of $\lambda$ was set to 0.2 during interpolation to obtain semi-factuals. For DiCE, we used the publicly available library\footnote{\url{https://github.com/interpretml/DiCE}} and customized the objective function to obtain semi-factuals. The classifier model was trained using a Random Forest (RF) Classifier\footnote{We used default parameters to train RF across the datasets. The classifier's key role is mostly to validate the class-membership of the generated semi-factuals.} and 3 diverse explanations were retrieved using the randomized search method.  In the Local Region method, the surrogate model was trained with a minimum of 200 instances from each class. We followed the original implementation of DSER\footnote{\url{https://github.com/HammerLabML/DiverseSemifactualsReject}} where we used a k-NN classifier to fit the conformal predictor. The hyperparameters including rejection threshold $\theta$ was identified using grid search cross-validation. For MDN method, we used the modified scoring function, $\textit{sfs}_{v2}()$. For continuous features, we selected the threshold of $\pm20\%$ of standard deviation to determine if they are same in the $\textit{sfs}_{v2}()$ function. Finally, for S-GEN we used the default implementation available\footnote{{\url{https://github.com/EoinKenny/Semifactual_Recourse_Generation}}} for causal and non-causal setting and retrieve 3 diverse semi-factual explanations (see GitHub for causal and non-causal datasets).

All the experiments were conducted in Python 3.9 environment on a Ubuntu 23.10 server with AMD EPYC 7443 24-Core Processor. The source code, data and results is available at {\url{https://github.com/itsaugat/sf_methods/}.

\subsection{Results \& Discussion}

Table 1 shows the normalized raw scores for the 8 methods across the 7 datasets on the 5 metrics.  Figure \ref{fig:mean_rank_2} summarises the results as mean ranks for each method (collapsing over datasets and metrics).  This graph immediately allows us to answer our main question, ``Are the best semi-factuals found by using counterfactuals as guides?".   The short answer is ``NO". If counterfactual guidance were critical then the counterfacual-guide methods (the grey bars) should be the top-ranked methods to the left, with the counterfactual-free methods (the black bars) to the right.  However, instead, the rank order alternates between the two groups of methods. Indeed, a counterfactual-free method, MDN appears to be the best. Furthermore, the top scores for each metric are distributed over methods of both groups (see bold-text scores in Table 1).  Hence, using counterfactuals as guides in semi-factual generation does not determine success on key metrics.  So, what is it then that makes one method better than another in these tests?  

Figure \ref{fig:results} shows the ranks for each method by metric (collapsing over dataset) with a separate radar chart for each group. In these charts it is clear that no method dominates coming first across most metrics.  Indeed, the top-3 methods -- MDN, C2C-VAE, Local-Region -- are really not that good on further analysis of their performance; many do well on one or two metrics but then poorly on others. 

\begin{wrapfigure}{r}{0.5\textwidth}
  \vspace{-10mm}
  \begin{center}
    \includegraphics[width=0.5\textwidth]{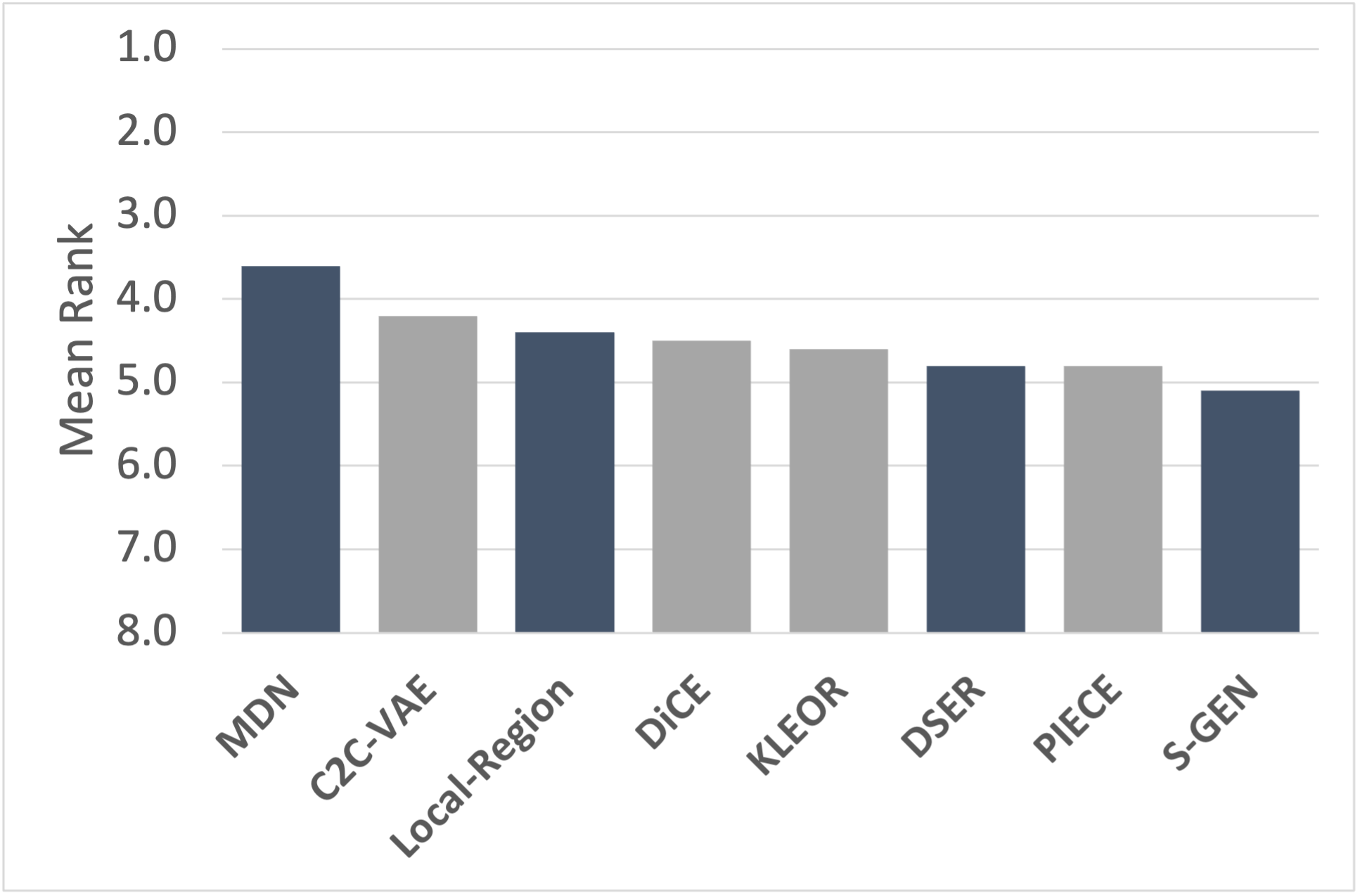}
  \end{center}
  \vspace{-0pt}
  \caption{ \small Mean Ranks of Counterfactual-Free (black) and Counterfactual-Guided (grey) Semi-Factual Methods}
  \vspace{-0pt}
  \label{fig:mean_rank_2}
    \vspace{-6mm}
\end{wrapfigure}

MDN does very well on Confusability (1st) and Plausibility (2nd) but is mediocre on Sparsity (4th), Distance (4th) and Robustness (5th). As MDN selects known instances in the query-class it is not surprising that its semi-factuals are plausible and in the class but they then tend not to be very distant from the query, to be less sparse and (sometimes) unstable outliers. 

In contrast, C2C-VAE does much better on Robustness (1st) and Distance (2nd) and but then is less good on Plausibility (5th) and poor on Confusability (7th) and Sparsity (8th).  C2C-VAE analyses the differences between instances with respect to classes, information that clearly enables the identification of stable semi-factuals distant from the query; but, these semi-factuals go too far, becoming confusable with counterfactuals and rely on too many feature-differences.   

Ironically, the older, third-placed Local-Region method could be said to be the best, if we were adopting a balanced-scorecard approach; arguably, it is moderately good across most metrics showing Distance (3rd), Plausibility (4th),  Confusability (4th), Robustness (5th) and Sparsity (7th). 

Other key insights can also be gained from these results. Semi-factuals obtained from the methods that select existing instances (MDN, KLEOR and Local-Region) are more plausible as compared to those produced by generative methods (DiCE, PIECE, S-GEN and C2C-VAE). Similarly, on a group-level, the counterfactual-guided methods (C2C-VAE, PIECE and DiCE) produces more robust semi-factuals than counterfactual-free methods. 

\begin{figure}[t]
\centering
\begin{subfigure}{.5\textwidth}
  \centering
  \includegraphics[width=.9\linewidth]{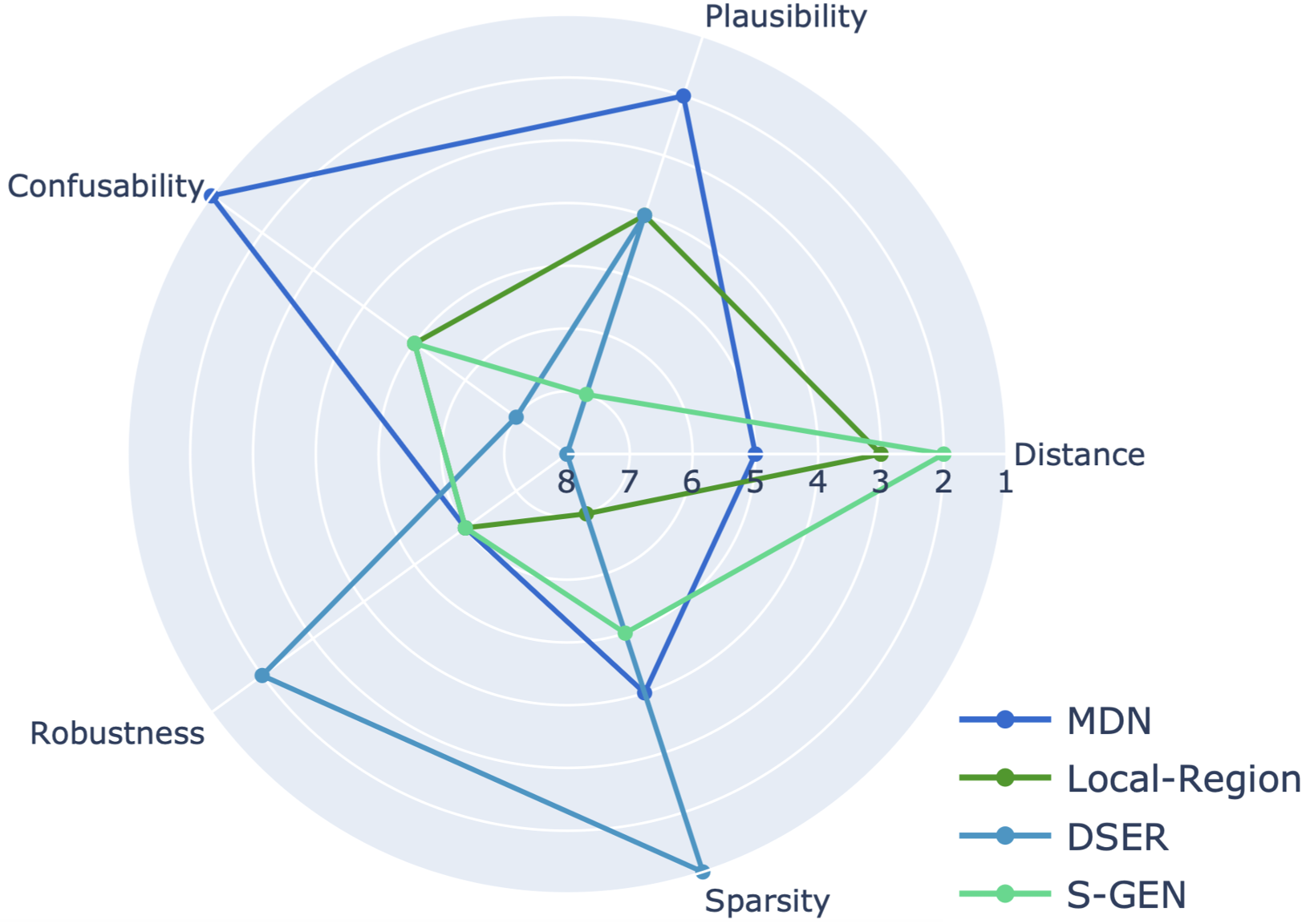}
  \caption{Counterfactual-Free Methods}
  \label{fig:sub3}
\end{subfigure}%
\begin{subfigure}{.5\textwidth}
  \centering
  \includegraphics[width=.9\linewidth]{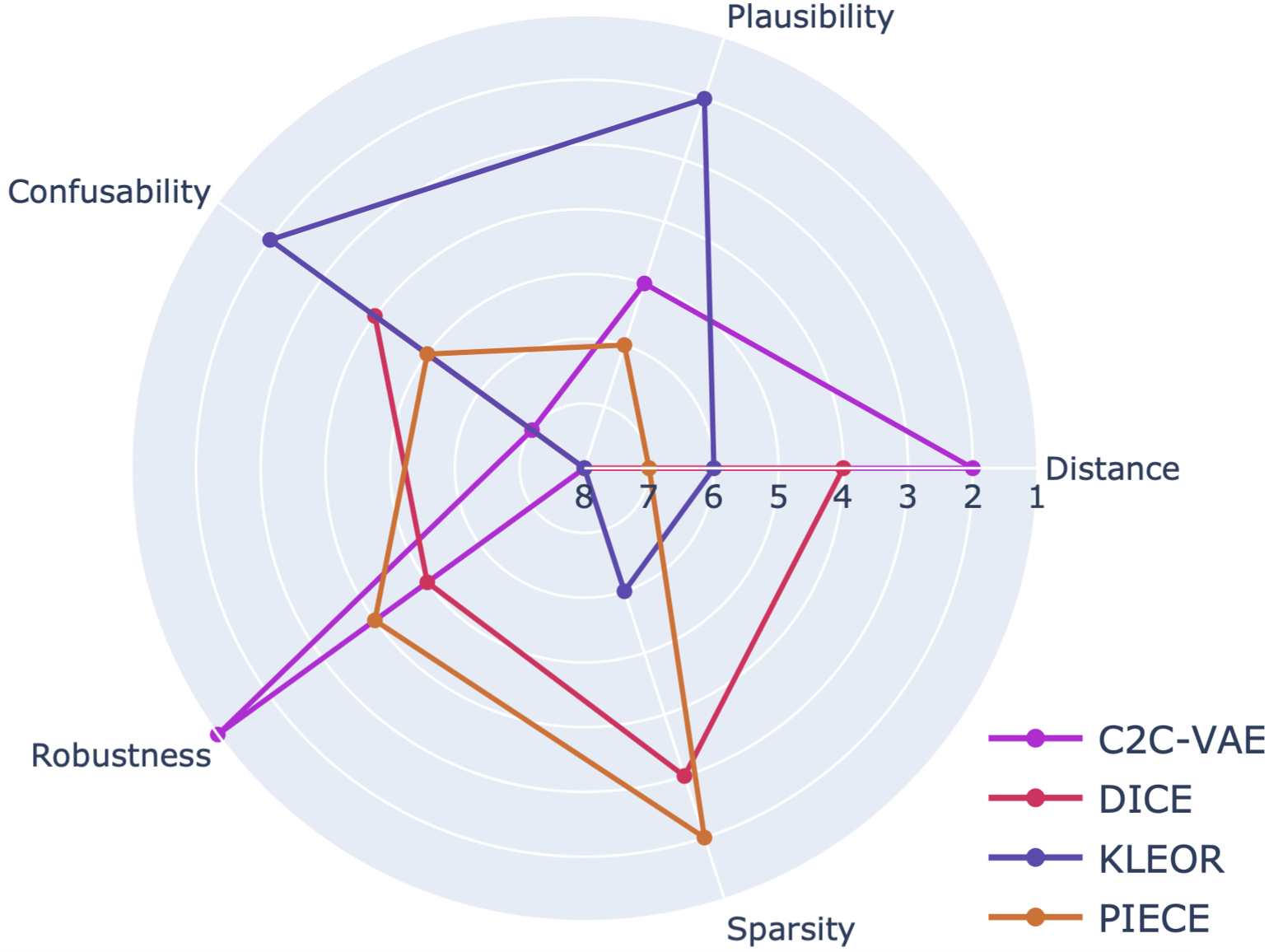}
  \caption{Counterfactual-Guided Methods}
  \label{fig:sub4}
\end{subfigure}
\caption{Median Ranks (across datasets) of Counterfactual-Free (left) and Counterfactual-Guided (right) methods for each measure. Points away from the center of graphs represent higher rank (better performance).}
\label{fig:results}
\end{figure}

\setlength{\tabcolsep}{6pt}
\begin{table}[!ht]
\caption{Scores for each method on the five metrics across different datasets (values are normalized between 0 and 1). The best score for a given dataset is shown in bold, with arrows after metrics showing direction of best scores.}
 \centering
\scriptsize
\begin{tabular}{@{}lllllllll@{}}
\toprule
Metrics & Method & D1 & D2 & D3 & D4 & D5 & D6 & D7                 
\\ \midrule
& MDN & 0.255 & 0.205 & 0.172 & 0.471 & 0.628 & 0.476 & 0.285   \\
& Local-Region & 0.364 & 0.305 & 0.217 & \textbf{0.65} & \textbf{0.984} & 0.506 & 0.152 
\\
& DSER & 0.024 & 0.027 & 0.04 & 0.026 & 0.042 & 0.038 & 0.035   
\\
Distance ($\uparrow$) & S-GEN & \textbf{0.518} & 0.313 & 0.488 & 0.5 & 0.463 & \textbf{0.95}  & 0.23
\\
& \textit{C2C\_VAE} & 0.488 & \textbf{0.352} & 0.36 & 0.451 & 0.762 & 0.586 & 0.361
\\
& \textit{DiCE} & 0.427 & 0.304 & \textbf{0.492} & 0.399 & 0.354 & 0.494 & \textbf{0.498} 
\\
& \textit{KLEOR} & 0.176 & 0.143 & 0.245 & 0.363 & 0.714 & 0.446 & 0.151  \\
& \textit{PIECE} & 0.124 & 0.2 & 0.459 & 0.141 & 0.076 & 0.194 & 0.135    
\\ \midrule
& MDN & \textbf{0.046} & 0.041 & 0.035 & 0.223 & \textbf{0.285} & 0.205 & 0.023 
\\
& Local-Region & 0.069 & 0.047 & 0.049 & 0.258 & 0.302 & 0.219 
& 0.037 
\\
& DSER & 0.062 & 0.052 & 0.085 & 0.267 & 0.319 & 0.263 & 0.045 
\\
Plausibility ($\downarrow$) & S-GEN & 0.126 & 0.104 & 0.209 & 0.261 & 0.39
& 0.489 & 0.038                     
\\
& \textit{C2C\_VAE} & 0.078 & 0.064 & 0.091 & \textbf{0.158} & 0.377 
& \textbf{0.169} & 0.024                     
\\
& \textit{DiCE} & 0.201 & 0.099 & 0.352 & 0.351 & 0.428 & 0.473 & 0.228  \\
& \textit{KLEOR} & 0.048 & \textbf{0.031} & \textbf{0.032} & 0.193 & 0.294 & 0.203 & \textbf{0.022}
\\
& \textit{PIECE} & 0.108 & 0.095 & 0.123 & 0.254 & 0.359 & 0.313 & 0.046
\\ \midrule
& MDN & $\textbf{0}$ & \textbf{0} & 0.419 & 0.321 & \textbf{0} & 0.134 & \textbf{0}  
\\
& Local-Region & 0.888 & 0.988 & 0.344 & 0.913 & 0.66 & 0.567	& 0.688
\\
& DSER & 1	& 0.99 & 0.925	& 0.991	& 1	& 0.9 & 1    
\\
Confusability ($\downarrow$)  & S-GEN & 0.9 & 0.88	& 0.95	& \textbf{0}	& 0.226 & 1 & 0.91
\\
& \textit{C2C\_VAE} & 0.975	& 0.99	& 0.957	& 1	& 0.901	& 0.867	& 0.869
\\
& \textit{DiCE} & 0.844	& 0.614	& 0.913	& 0.758	& 0.595	& 0.634	& 0.915  \\
& \textit{KLEOR} & 0.227	& 0.318	& \textbf{0}	& 0.525	& 0.033	& \textbf{0}	& 0.014 
\\
& \textit{PIECE} & 0.796	& 1	& 1	& 0.719	& 0.611	& 0.867	& 0.978
\\ \midrule
& MDN & 0.996 & 0.866 & 0.9994 & 0.548	& 0.991	& 0.748	& 0.905
\\
& Local-Region & 0.967	& 0.908	& 0.983	& \textbf{1}	& 0.889	& 0.873	& 0.942
\\
& DSER & 0.999	& \textbf{0.99}	& \textbf{1}	& 0.984	& 0.998	& 0.998	& 0.995
\\
Robustness ($\uparrow$) & S-GEN & 0.991 & 0.982	& 0	& 0.785	& 0.214	& 0.985	& 0.626
\\
& \textit{C2C\_VAE} & \textbf{1}	& 0.975	& \textbf{1} & \textbf{1}	&\textbf{1}	& \textbf{1} & \textbf{1}
\\
& \textit{DiCE} & 0.975	& 0.894	& 0.997	& 0.896	& 0.983	& 0.991	& 0.922
\\
& \textit{KLEOR} & 0	& 0	& 0.753	& 0	& 0	& 0	& 0
\\
& \textit{PIECE} & 0.949	& 0.988	& 0.967	& 0.953	& 0.983	& 0.991	& 0.939
\\ \midrule
& MDN & 0.53 & 0.43	& 0.08	& 0.14	& 0.17	& 0.09	& 0.35
\\
& Local-Region & 0.19 & 0.29 & 0.06	& 0.13	& 0.08	& 0.06	
&0.17
\\
& DSER & \textbf{0.99} & \textbf{0.95} & \textbf{0.98}	& \textbf{0.95}	& \textbf{0.9}	& \textbf{0.94}	& 0.85
\\
Sparsity ($\uparrow$) & S-GEN & 0.33	& 0.52	& 0.06	& 0.16	& 0.15	& 0.07	& 0.21
\\
& \textit{C2C\_VAE} & 0.1 & 0.25	& 0.05	& 0.125	& 0.07	& 0.05	& 0.13
\\
& \textit{DiCE} & 0.7 & 0.81	& 0.74	& 0.38	& 0.65	& 0.74	& 0.7
\\
& \textit{KLEOR} & 0.42	& 0.37	& 0.06	& 0.12	& 0.13	& 0.06	& 0.26
\\
& \textit{PIECE} & 0.783	& 0.81	& 0.98	& 0.93	& 0.608	& 0.84	& \textbf{0.96}
\\ \bottomrule
\end{tabular}
\end{table}

\section{Conclusions}
In XAI literature, semi-factual explanation methods have been studied from two perspectives: counterfactual-free and counterfactual-guided approaches. This paper addressed the question of whether relying on counterfactuals as guides necessarily and always produces the best semi-factuals. We find that methods from both approaches show good results on some evaluation metrics, while performing poorly on others. As such, the use of counterfactual guides does not appear to be a major determining factor in finding the best semi-factuals.  Indeed, it is hard to escape the conclusion that other factors better guide the process; for instance, relying on known data-points (as in MDN and KLEOR) rather than on perturbed instances or using the difference-space (as in C2C-VAE). Hence, the most promising direction for future research may be to extract what is best from the current methods and somehow combine them in one.

The current work has a number of limitations that should be mentioned. This work primarily focuses on tabular data; so, it would be interesting to determine whether the results extend to other data-types (e.g., images and time series). We also badly need more user studies. In many respects, we do not know whether people find the semi-factuals produced by any of these methods are psychologically valid. Unfortunately, the few users studies that have been done suffer from design flaws (see \cite{kenny2023sf,nugent2009gaining,doyle2004explanation}). However, what is clear, is that there is a lot more to do on the computation of semi-factual explanations before we can be confident about using them in deployed systems. 

\hfill\break
\noindent \textbf{Acknowledgements.} This research was supported by Science Foundation Ireland via the Insight SFI Research Centre for Data Analytics (12/RC/2289-P2).  For the purpose of Open Access, the author has applied a CC BY copyright to any Author Accepted Manuscript version arising from this submission.

\clearpage
\bibliographystyle{splncs04}
\bibliography{bibliography}

\end{document}